\pdfoutput=1

\documentclass[11pt]{article}

\usepackage[table]{xcolor}

\usepackage[final]{acl}

\usepackage{times}
\usepackage{latexsym}

\usepackage[T1]{fontenc}

\usepackage[utf8]{inputenc}

\usepackage{microtype}

\usepackage{inconsolata}

\usepackage{graphicx}

\usepackage{multirow}
\usepackage{array}
\newcommand{\ph}[1]{\phantom{#1}}
\usepackage{quoting}
\quotingsetup{leftmargin=10pt, rightmargin=0pt}
\usepackage{caption} 

\title{Why Do We Laugh? Annotation and Taxonomy Generation\\for Laughable Contexts in Spontaneous Text Conversation}

\author{
 \textbf{Koji Inoue},
 \textbf{Mikey Elmers},
 \textbf{Divesh Lala},
 \textbf{Tatsuya Kawahara}
\\
\\
 Graduate School of Informatics, Kyoto University, Japan, \\
\\
 \small{
   \textbf{Correspondence:} \href{mailto:inoue@sap.ist.i.kyoto-u.ac.jp}{inoue@sap.ist.i.kyoto-u.ac.jp}
 }
}

\begin{document}
\maketitle
\begin{abstract}
Laughter serves as a multifaceted communicative signal in human interaction, yet its identification within dialogue presents a significant challenge for conversational AI systems.
This study addresses this challenge by annotating laughable contexts in Japanese spontaneous text conversation data and developing a taxonomy to classify the underlying reasons for such contexts.
Initially, multiple annotators manually labeled laughable contexts using a binary decision (laughable or non-laughable).
Subsequently, an LLM was used to generate explanations for the binary annotations of laughable contexts, which were then categorized into a taxonomy comprising ten categories, including ``Empathy and Affinity'' and ``Humor and Surprise,'' highlighting the diverse range of laughter-inducing scenarios.
The study also evaluated GPT-4o's performance in recognizing the majority labels of laughable contexts, achieving an F1 score of 43.14\%.
These findings contribute to the advancement of conversational AI by establishing a foundation for more nuanced recognition and generation of laughter, ultimately fostering more natural and engaging human-AI interactions.
\end{abstract}

\renewcommand{\thefootnote}{\fnsymbol{footnote}}
\footnote[0]{This paper has been accepted for presentation at International Workshop on Spoken Dialogue Systems Technology 2025 (IWSDS 2025) and represents the author’s version of the work.}
\renewcommand{\thefootnote}{\arabic{footnote}}

\section{Introduction}

In human dialogue, laughter serves as a communicative signal conveying humor, empathy, surprise, or social bonding~\cite{norrick1993conversational,glenn2003laughter,attardo2009linguistic}.
However, its mechanisms are complex and multifaceted, and understanding them remains a long-term challenge for dialogue systems aiming to achieve human-like interaction~\cite{tian2016we,turker2017analysis,mazzocconi2020s,inoue2022shared,ludusan-wagner-2023-effect,perkins2024linguistic}.
Furthermore, traditional approaches to modeling laughter and humor have often been limited to scenarios involving explicit auditory or visual stimuli, with few addressing the subtle contextual nuances present in spontaneous dialogue~\cite{bertero2016long,choube2020punchline,jentzsch-kersting-2023-chatgpt,ko-etal-2023-language,hessel-etal-2023-androids}.
Therefore, elucidating the underlying reasons for laughter in spontaneous dialogue data can contribute to making large language model (LLM)-based dialogue more natural and empathetic.
However, annotating the reasons for laughter in any formalized manner has been prohibitively time- and labor-intensive, leaving the field largely reliant on qualitative approaches through conversational analysis.

\begin{table}[t]
    \centering
    \rowcolors{1}{white}{gray!20}
    \renewcommand{\arraystretch}{1.2}
    {\small
    \caption{Annotation example for laughable context (majority voting, translated from Japanese)}
    \label{tab:annotation}
    \begin{tabular}{c m{45mm} c}
    \hline
     & \multicolumn{1}{c}{Utterance} & Laughable? \\
    \hline
    A: & I think that’s a wonderful attitude. I always end up talking about myself, so I should follow your example. & NO
    \\
    B: & Is that so? But does your husband listen to your stories? & NO \\
    A: & Yes, yes, he listens to me. I wonder if I’m putting too much on him? & NO \\
    B: & I don’t think so! He’s so kind. My husband doesn’t seem to listen to me. Huh, that’s strange. & \textbf{YES} \\
    \hline
    \end{tabular}
    }
\end{table}

In this study, we address the question of ``why do we laugh?'' from an informatics perspective by proposing a semi-automated approach to constructing taxonomy labels for the reasons of laughter.
First, to identify target segments, multiple annotators were asked to perform a simple binary classification on each utterance in dialogue data, determining whether it was ``laughable'' or not, as shown in Table~\ref{tab:annotation}.
Subsequently, for contexts labeled as ``laughable'' based on the majority voting, we used an LLM (GPT-4o) to generate the reasoning sentence behind this judgment and further classified these generated reasons into distinct categories (taxonomy labels).
This semi-automated taxonomy generation approach is generalizable and can be particularly effective in scenarios where manual annotation is limited to simpler labels, such as emotion labeling.

The purpose of this research is to contribute toward more nuanced conversational AI systems that can recognize and even anticipate moments for laughter, ultimately fostering more natural interactions between humans and machines.
Ideally, such systems should be able to respond with the correct acoustics, delay, and consider group size for different laughter types~\cite{truong12binterspeech}.
Our findings reveal that AI can improve our understanding of laughter and offer a foundation for future research in AI context-sensitive recognition.

\begin{table}[t]
    \centering
    \setlength{\tabcolsep}{3mm}
    \caption{Number of samples in each ratio of annotators judged as laughable (laughable agreement)}
    \label{tab:annotation:dist}
    { 
    \begin{tabular}{cc}
    \hline
    Laughable agreement & \multicolumn{1}{c}{\# sample} \\
    \hline
    1.0 ~ (5/5) & \ph{00}163 ~ (\ph{0}0.64\%) \\
    0.8 ~ (4/5) & \ph{00}845 ~ (\ph{0}3.34\%) \\
    0.6 ~ (3/5) & \ph{0}2731 ~ (10.80\%) \\
    0.4 ~ (2/5) & \ph{0}8143 ~ (32.20\%) \\
    0.2 ~ (1/5) & 11928 ~ (47.17\%) \\
    0.0 ~ (0/5) & \ph{0}1479 ~ (\ph{0}5.85\%) \\
    \hline
    \end{tabular}
    }
\end{table}

\section{Annotation of Laughable Context}

We annotated laughable contexts in the RealPersonaChat dataset~\cite{yamashita2023realpersonachat}.
This textual data contains one-on-one Japanese spontaneous conversation where participants chat without assuming assigned personas.
It includes approximately 30 utterances per conversation, totaling around 14,000 dialogues.
We annotated 900 dialogues, with plans to annotate the remainder in future work.

During the annotation process, each annotator reviewed each dialogue and, after the initial two greeting utterances, made a binary decision for whether the next person would laugh (\textit{laughable}) or not. 
Five annotators assigned these binary labels to each utterance.
Table~\ref{tab:annotation:dist} summarizes the agreement amongst annotators for laughable labels, which we refer to as ``laughable agreement''.
While some samples showed clear agreement (either all or none of the annotators marked them as laughable), there were also numerous split samples, highlighting the subjectivity and complexity of the task.
If we applied a majority voting process, 3,739 contexts (14.8\%) were labeled as laughable, and 21,550 contexts (85.2\%) as non-laughable.

Table~\ref{tab:annotation} illustrates a laughable context example.
In this dialogue, person A’s final utterance is self-contradictory, requiring high-level comprehension of the dialogue context.
These annotations underscore the significance of cultural context and conversational flow in interpreting laughter cues.


\begin{table*}[t]
    \centering
    \rowcolors{1}{white}{gray!20}
    \setlength{\tabcolsep}{1mm}
    \renewcommand{\arraystretch}{1.2}
    \caption{Generated taxonomy labels for laughable context reasoning, number of samples assigned to each taxonomy label, and related references for each taxonomy label}
    {\footnotesize
    \label{tab:taxonomy:explanation}
    \begin{tabular}{cm{25mm}m{88mm}>{\centering\arraybackslash}m{10mm}m{20mm}}
    \hline
    & \multicolumn{1}{c}{Label name} & \multicolumn{1}{c}{Explanation} & \#sample & \multicolumn{1}{c}{Reference} \\
    \hline
    (1) & Empathy and Affinity & Situations where a sense of closeness and laughter is generated by sharing common experiences or emotions in a conversation. This includes empathy for shared hobbies or everyday relatable situations. & 3013 (80.6\%) & \cite{hay2001pragmatics,garbarski2016interviewing} \\
    (2) & Humor and Surprise & Cases where humor or an element of surprise in the statement triggers laughter. This includes unexpected twists, wordplay, and exaggeration. & 3233 (86.5\%) & \cite{dynel2009beyond,martin2018psychology} \\
    (3) & Relaxed Atmosphere & Situations where the conversation progresses in a calm, relaxed atmosphere, naturally leading to laughter. Lighthearted exchanges and conversations with jokes fall into this category. & 2955 (79.0\%) & \cite{vettin2004laughter} \\
    (4) & Self-Disclosure and Friendliness & Situations where sharing personal stories or past mistakes creates a sense of approachability and triggers laughter. Self-disclosure that makes the other person feel at ease is also included. & 475 (12.7\%) & \cite{gelkopf1996humor} \\ 
    (5) & Cultural Background and Shared Understanding & Laughter based on specific cultural backgrounds or shared understandings. This includes jokes related to a particular region or culture or remarks based on common superstitions or folklore. & 176 (4.7\%) & \cite{bryant2022laughter,kamilouglu2022perception} \\
    (6) & Nostalgia and Fondness & Situations where past memories or nostalgic topics trigger laughter. This includes shared past experiences and the enjoyment of recalling familiar events. & 204 (5.5\%) & \cite{bazzini2007effect} \\ 
    (7) & Self-Deprecating Humor & Situations where turning one’s flaws or mistakes into humor makes the other person feel closer and triggers laughter. Slightly putting oneself down can also give the other person a sense of ease. & 404 (10.8\%) & \cite{atkinson2015self} \\
    (8) & Defying Expectations & Situations where intentionally defying the flow of conversation or the other person’s expectations creates an element of surprise and triggers laughter. This includes unexpected responses or developments. & 323 (8.6\%) & \cite{ginzburg2020laughter,xu2022analysis} \\
    (9) & Positive Energy & Situations where positive emotions or energy in the conversation bring a smile to the other person. Enjoyable topics and positive comments fall into this category. & 338 (9.0\%) & \cite{wang2024multimodal} \\
    (10) & Exaggeration & Situations where exaggerating things gives a comical impression and triggers laughter. Exaggerated expressions can be amusing to the listener. & 478 (12.8\%) & \cite{mccarthy2004there} \\
    \hline
    \end{tabular}
    }
\end{table*}

\section{Generating Taxonomy of Laughable Reason} \label{sec:taxonomy}

Towards developing human-like laughter behaviors in LLMs, we investigated the reasons behind human annotators' recognition of laughable contexts.
In this section, we used only samples with majority labels marked as laughable (3,739 samples).
Since manual annotations are costly, we utilized GPT-4o to generate explanations for the human judgments.
For example, a generated reason for the example context in Table~\ref{tab:annotation} is:
\begin{quoting}
{\it
If we were to speculate on the reasons a third party might judge that Person A laughed at Person B's final remark in this conversation, the following points can be considered:

Element of Humor: Person B's comment, ``My husband doesn’t seem to listen to me. Huh, that’s strange.,'' contains a touch of self-deprecating humor. This lighthearted tone, making fun of their own situation, can be amusing to the listener. (...)



Relaxed Atmosphere: The overall tone of the conversation seems light and relaxed, and Person B's comment might have been perceived as a playful joke in line with this mood.

A combination of these factors may have led the third party to interpret that Person A laughed in response to Person B's remark.
}
\end{quoting}

\begin{table*}
    \centering
    \setlength{\tabcolsep}{2mm}
    \renewcommand{\arraystretch}{1.2}
    \caption{Laughable context recognition performance of GPT-4o on each taxonomy label}
    {\footnotesize
    \label{tab:taxonomy:label}
    \begin{tabular}{cm{65mm}cc}
    \hline
    & \multicolumn{1}{c}{\multirow{2}{*}{Label name}} & \multicolumn{2}{c}{\# output by GPT-4o} \\
    \cline{3-4}
    & & Laughable (correct) & Non-laughable (incorrect) \\
    \hline
    \rowcolor{gray!20} (1) & Empathy and Affinity & 1226 ~ (40.69\%) & 1787 ~ (59.31\%) \\
    (2) & Humor and Surprise & 1571 ~ (48.59\%) & 1662 ~ (51.41\%)  \\
    \rowcolor{gray!20} (3) & Relaxed Atmosphere & 1257 ~ (42.54\%) & 1698 ~ (57.46\%)  \\
    (4) & Self-Disclosure and Friendliness & \ph{0}232 ~ (48.84\%) & \ph{0}243 ~ (51.16\%)  \\
    \rowcolor{gray!20} (5) & Cultural Background and Shared Understanding & \ph{0}102 ~ (57.95\%) & \ph{00}74 ~ (42.05\%) \\
    (6) & Nostalgia and Fondness & \ph{00}62 ~ (30.39\%) & \ph{0}142 ~ (69.61\%)  \\
    \rowcolor{gray!20} (7) & Self-Deprecating Humor & \ph{0}255 ~ (63.12\%) & \ph{0}149 ~ (36.88\%)  \\
    (8) & Defying Expectations & \ph{0}227 ~ (70.28\%) & \ph{00}96 ~ (29.72\%)  \\
    \rowcolor{gray!20} (9) & Positive Energy & \ph{00}50 ~ (14.79\%) & \ph{0}288 ~ (85.21\%)  \\
    (10) & Exaggeration & \ph{0}239 ~ (50.00\%) & \ph{0}239 ~ (50.00\%) \\
    \hline
    \end{tabular}
    }
\end{table*}

\begin{table}[t]
    \centering
    \setlength{\tabcolsep}{1mm}
    \rowcolors{1}{white}{gray!20}
    \renewcommand{\arraystretch}{1.2}
    \caption{Example context for ``Nostalgia and Fondness'' (translated from Japanese)}
    \label{tab:taxonomy:example1}
    {\small
    \begin{tabular}{c m{70mm}}
    \hline
     & \multicolumn{1}{c}{Utterance} \\
    \hline
    A: & Do you also consume milk or yogurt for calcium? \\
    B: & I drink milk with Milo in it. I also eat yogurt as a snack. \\
    A: & That’s really well-balanced! \\
    B: & Yes, health is important. \\
    A: & It’s been a while since I last heard about Milo. \\
    \hline
    \end{tabular}
    }
\end{table}

\begin{table}[t]
    \centering
    \setlength{\tabcolsep}{1mm}
    \rowcolors{1}{white}{gray!20}
    \renewcommand{\arraystretch}{1.2}
    \caption{Example context for ``Positive Energy'' (translated from Japanese)}
    \label{tab:taxonomy:example2}
    {\small
    \begin{tabular}{c m{70mm}}
    \hline
     & \multicolumn{1}{c}{Utterance} \\
    \hline
    A: & Oh, as they grow up, that kind of help really makes a difference, doesn’t it? \\
    B: & Absolutely! It’s such a joy, isn’t it? So reassuring. \\
    A: & When they’re little, it’s like a never-ending story of challenges, isn’t it? \\
    B: & Haha, so true. All we have now are funny memories of those times. \\
    A: & Once you get through it, those challenges become stories you can laugh about, and you feel glad you went through them. \\
    \hline
    \end{tabular}
    }
\end{table}

We then aimed to summarize the generated reasoning texts for laughable contexts by applying a taxonomy generation approach using LLMs~\cite{wan2024tnt}. First, we randomly divided the generated reason samples into smaller subsets, each comprising roughly 5\% of the samples.
Starting with the first subset, we gave the reason sentences GPT-4o to generate initial taxonomy labels and those explanations, which we manually validated when necessary.
We then iteratively refined the taxonomy by having the LLM update it based on the previous taxonomy and the reason sentence data from each new subset, continuing this process until all data were processed.
This resulted in ten taxonomy labels, summarized in Table~\ref{tab:taxonomy:explanation}, including categories such as (1) Empathy and Affinity and (2) Humor and Surprise.

After generating these taxonomy labels, we used the LLM to assign them to each reason sample, allowing for multiple labels per sample.
The labeling results are shown on the right side of Table~\ref{tab:taxonomy:explanation}.
While some categories, such as (1) Empathy and Affinity, were predominant, many samples were also assigned to other categories, including (4) Self-Disclosure and Friendliness and (5) Cultural Background and Shared Understanding.
This broad distribution across categories reinforces the validity of the generated taxonomy.
A correlation matrix showing relationships between the taxonomy labels is provided in Appendix~\ref{sec:appendix}.
Finally, we reviewed related studies in conversational analysis, as listed on the right side of Table~\ref{tab:taxonomy:explanation}.
These studies further substantiate the explanatory power of our taxonomy within the context of conversational analysis research.

\section{LLM's Performance on Laughable Context Recognition}

We then examined how much LLMs, specifically GPT-4o, can recognize the laughable contexts in spontaneous text conversation.
The model was tested in a zero-shot setting, instructed to first analyze the conversational context and then determine its laughability as a binary.
The provided prompt included a task description for laughable context recognition, followed by a Chain-of-Thought (CoT) reasoning approach to encourage the model to consider the reasoning behind its decision step by step.
We evaluated GPT-4o’s performance against the majority labels, achieving an F1 score of 43.14\%, with a precision of 41.66\% and recall of 44.72\%.
While this score was significantly above the chance level (14.8\%), capturing the nuanced subtleties of conversational humor remains challenging.

We then further examined the LLM’s performance on each generated taxonomy label.
Table~\ref{tab:taxonomy:label} shows the distribution of binary outputs by GPT-4o and its accuracy within each label. 
First, the primary labels, from (1) to (3), showed similar accuracy rates, ranging from 40\% to 50\%.
Additionally, we observed comparatively higher scores for (5) \textit{Cultural Background and Shared Understanding}, (7) \textit{Self-Deprecating Humor}, and (8) \textit{Defying Expectations}, suggesting that the current LLM may effectively capture these contexts.
In contrast, categories like (6) \textit{Nostalgia and Fondness} and (9) \textit{Positive Energy} displayed lower accuracy, potentially highlighting limitations in the LLM’s understanding.

Table~\ref{tab:taxonomy:example1} presents an example dialogue context where the LLM marked non-laughable for the final utterance from person A, despite a positive majority label with a (6) \textit{Nostalgia and Fondness} reason.
This context was also assigned the (2) \textit{Humor and Surprise} and (3) \textit{Relaxed Atmosphere} labels.
In this example, the participants discuss a nostalgic memory of drinking a powdered beverage with milk.
The last utterance evokes nostalgia, implicitly inviting laughter.
Here, capturing person A’s sentiment seems to be difficult for the current LLM, but is essential for appropriate laughter response.

Table~\ref{tab:taxonomy:example2} provides an example for (9) \textit{Positive Energy} label.
This context was also assigned the (1) \textit{Empathy and Affinity} and (2) \textit{Humor and Surprise} labels.
The participants discussed a challenging experience with childcare, but in the final utterance, person A reflects positively on the experience after some time has passed.
Although the story itself recounts a difficult time, it is now viewed positively, making it laughable.
This example suggests that the LLM needs to comprehend the temporal structure of the story and the person’s current feelings to accurately interpret the context as laughable.

\section{Conclusion}

This study investigated laughter in the context of conversational AI by annotating laughable contexts within a Japanese text dialogue dataset.
A taxonomy of ten distinct reasons for laughter was generated by an LLM, providing valuable insights into the multifaceted nature of laughter.
Subsequently, this study evaluated the ability of GPT-4o to recognize those laughable contexts.
While the model's performance surpassed chance levels, it highlighted the inherent challenges in capturing the nuances of conversational humor.

This automated approach employed for reasoning and taxonomy generation with LLMs can be applied in other scenarios where only binary (or simplified) decision labels from human annotators are available, yet more fine-grained explanations are required.
Future work will focus on expanding the dataset to cover other languages and cultural contexts, validating the generated taxonomy by incorporating additional linguistic research perspectives, exploring multimodal approaches, and including spoken dialogue to enhance AI's understanding of humor and social interaction.

\section*{Acknowledgement}

This work was supported by JST PREST JPMJPR24I4 and JSPS KAKENHI JP23K16901.

\bibliography{acl_latex}

\appendix

\section{Correlation Among Taxonomy Labels} \label{sec:appendix}

\begin{figure}[t]
    \centering
    \includegraphics[width=1\linewidth]{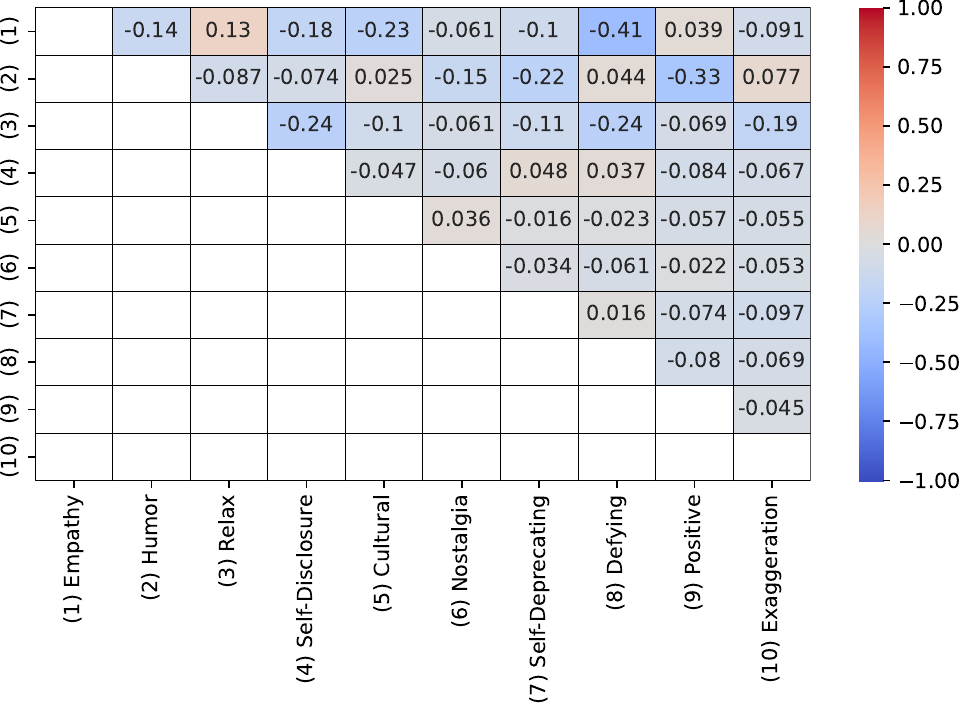}
    \caption{Correlation matrix of assigned taxonomy labels}
    \label{fig:correlation}
\end{figure}

Figure~\ref{fig:correlation} presents the correlation matrix of the assigned labels discussed in Section~\ref{sec:taxonomy}, where multiple labels can be assigned to the same laughable context.
For instance, ``Empathy and Affinity'' shows a weak positive correlation with ``Relaxed Atmosphere.''
Conversely, ``Empathy and Affinity'' exhibits a negative correlation with ``Defying Expressions.''
We also find a negative correlation between ``Humor and Surprise'' and ``Positive Energy,'' despite both being associated with positive sentiment.
This may be attributed to different expressive styles, with the former implicit and the latter explicit.
To gain deeper insight into the relationships between these labels, further qualitative analysis will be conducted in future work.

\end{document}